\documentclass{article}
\usepackage[dvipsnames]{xcolor}
\usepackage[final]{corl_2026}
\usepackage{bbm}
\usepackage{amsmath}
\usepackage{graphics}
\usepackage{graphicx}
\usepackage{multirow}
\usepackage{amssymb}
\usepackage{booktabs}
\usepackage{tikz}
\usepackage{xspace}
\usepackage{array}
\usepackage{colortbl}
\usepackage{caption}
\usepackage{placeins}
\usepackage[flushleft]{threeparttable}
\usepackage{subcaption}
\usepackage{algorithm}
\usepackage{algorithmic}
\usepackage{bm}
\usepackage{pgfplots}
\pgfplotsset{compat=1.18}
\usepackage{etoolbox}
\newif\ifshowrebuttalchanges
\showrebuttalchangesfalse
\newcommand{\rebuttalcolor}{%
  \ifshowrebuttalchanges
    \color{red}%
    \hypersetup{linkcolor=red,citecolor=red,urlcolor=red}%
  \fi
}
\DeclareRobustCommand{\rebuttalrev}[1]{{\rebuttalcolor#1}}
\definecolor{Green4}{RGB}{1,50,32}

\newif\ifshowcomments
\showcommentsfalse
\ifshowcomments
  \newcommand{\note}[3]{{\textcolor{#2}{[#1: #3]}}}
  \newcommand{\HZ}[1]{\note{Hui}{orange}{#1}}
  \newcommand{\ZJ}[1]{\note{Zijian}{blue}{#1}}
  \newcommand{\JS}[1]{\note{Jie}{magenta}{#1}}
  \newcommand{\TODO}[1]{\note{TODO}{red}{#1}}
\else
  \newcommand{\note}[3]{}
  \newcommand{\HZ}[1]{}
  \newcommand{\ZJ}[1]{}
  \newcommand{\JS}[1]{}
  \newcommand{\TODO}[1]{}
\fi

\newcommand{\methodname}{HiFun\xspace}

\title{\methodname: a Hierarchical Framework for Efficient Functional Dexterous Manipulation Learning}
\author{
  \normalfont
  \textbf{Linyi Huang}$^{1}$ \quad \textbf{Guowei Huai}$^{1}$ \quad
  \textbf{Weibin Liu}$^{1}$ \quad \textbf{Shulong Jiang}$^{1}$ \\
  \textbf{Ping Tan}$^{2}$ \quad \textbf{Weixuan Zhang}$^{1}$ \quad
  \textbf{Hui Zhang}$^{3}$ \quad \textbf{Jie Song}$^{1,2}$ \\
  $^{1}$HKUST (Guangzhou), China \quad $^{2}$HKUST, Hong Kong (China) \\
  $^{3}$ETH Zurich, Switzerland
}
\hypersetup{
  pdftitle={HiFun: a Hierarchical Framework for Efficient Functional Dexterous Manipulation Learning},
  pdfauthor={Linyi Huang, Guowei Huai, Weibin Liu, Shulong Jiang, Ping Tan, Weixuan Zhang, Hui Zhang, Jie Song}
}

\begin{document}
\maketitle

\begin{figure}[h]
\vspace{-4mm}
  \centering
  \includegraphics[width=\linewidth]{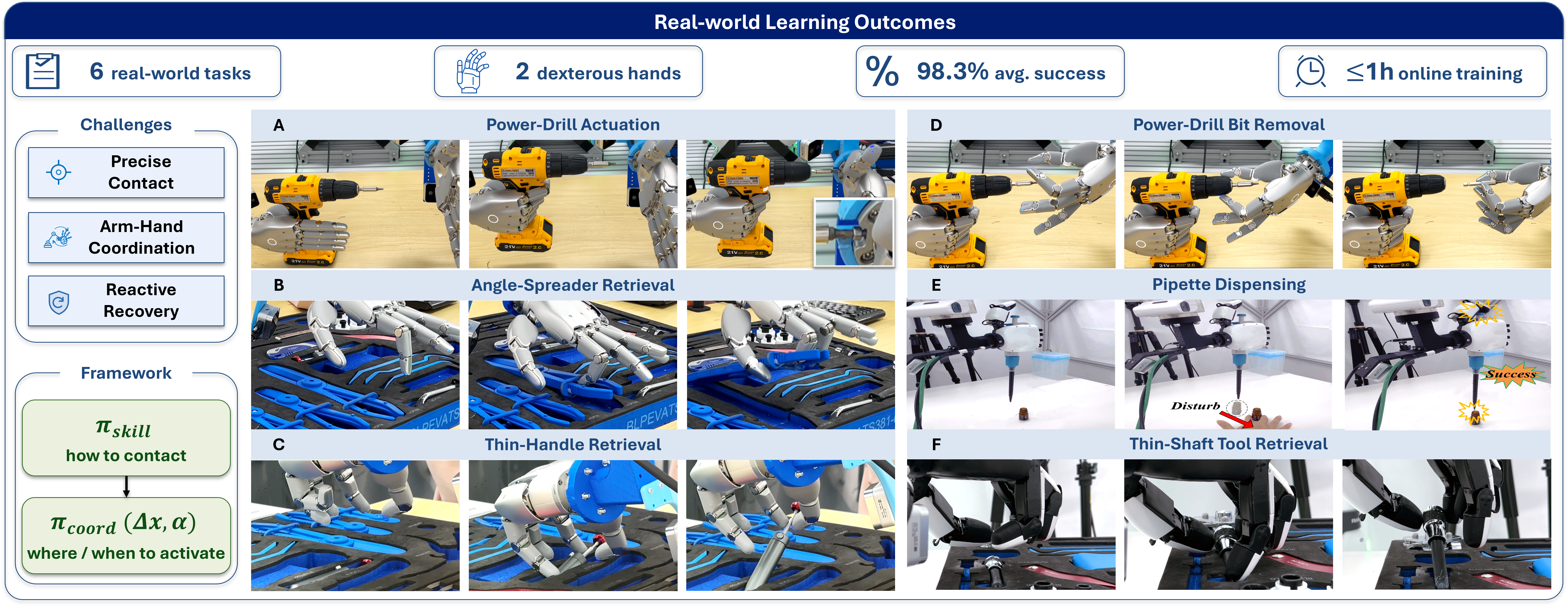}
  \captionof{figure}{\textbf{\methodname{} is a hierarchical framework for functional dexterous manipulation.} It learns contact-aware skills and arm-hand coordination across precise tool-use (A,E), contact-rich rotational manipulation (D), and constrained-space manipulation (B,C,F), achieving 98.3\% average success on six real-world tasks with two dexterous hands within one hour of online training.}
  \label{fig:teaser}
  \vspace{-2mm}
\end{figure}

\begin{abstract}
While multi-fingered dexterous robot hands have the potential to provide human-level dexterity, using them for functional manipulation remains difficult because task success depends on precise contact, coordinated arm-hand motion, and reactive recovery from execution errors.
Precise contact depends on complex contact dynamics that are difficult to model accurately in simulation, motivating real-world learning.
However, real-world demonstrations in the full arm-hand control space are costly and noisy because each trajectory must coordinate arm motion with high-DoF multi-finger contact.
Small execution deviations can further cause slipped contact or premature force application, making recovery from off-nominal contact states important for robust execution.
To deal with these challenges, we propose \methodname{}, a hierarchical real-world RL framework for functional dexterous manipulation tasks.
Our key insight is to decouple the control complexity into a functional hand skill for precise contact execution and an arm-hand coordination policy for arm control and hand-skill activation.
This design enables effective real-time human intervention for reactive recovery learning and efficient data collection with Dynamic Movement Primitive (DMP) augmentation.
Across six real-world tasks on two dexterous hands, \methodname{} achieves 98.3\% average success within one hour of online training and remains robust to dynamic disturbances.
\rebuttalrev{The project page is available at \url{https://hly-123.github.io/HiFun/}.}

\end{abstract}


\section{Introduction}
\vspace{-2mm}
Functional dexterous manipulation is central for robot applications from factories to homes.
We study this setting through representative tasks in precise tool use, contact-rich rotation, and constrained-space manipulation.
For example, Thin-Handle Tool Retrieval (Fig.~\ref{fig:teaser}C) requires extracting a foam-embedded handle through a narrow opening where direct grasping is blocked.
The robot must place fingertip contact on the exposed handle, coordinate arm motion with multi-finger levering to create access, and form a precision grasp without losing contact.
Small contact errors can cause slipping, trapping, or collisions with the opening.
These stages require precise contact, high-DoF arm-hand coordination, and recovery from contact errors, making functional dexterous manipulation challenging to learn and execute.

Specifically, precise contact requires the robot to place fingertips on task-relevant regions and regulate force as local contact changes.
High-DoF arm-hand coordination requires arm motion, multi-finger contact, and contact timing to be synchronized within one execution sequence.
Because fine-grained contact tasks are sensitive to small deviations, policies must recover from slipped contact, missed insertion, partial progress, or premature force application rather than only reproduce nominal successful trajectories.

Existing learning routes have made substantial progress, but these challenges remain only partially addressed for functional dexterity.
Precise contact can be explored at scale in simulation, enabling robust dexterous manipulation performance
\cite{openai2019learningdexterousinhandmanipulation, Chen2023VisualDexterity, zhang2024artigrasp, qi2025simple, zhang2024graspxl, pei2026assemblingpartshand}. However, sim-to-real transfer is difficult for functional manipulation due to its intricate contact dynamics. 
High-DoF arm-hand behaviors can be represented by imitation learning (IL) and pretrained visuomotor policies from demonstrations and robot datasets~\cite{chi2025diffusion, zhaoLearningFineGrainedBimanual2023, brohan2023rt2visionlanguageactionmodelstransfer, openxembodiment2023, kim2024openvlaopensourcevisionlanguageactionmodel, black2026pi0visionlanguageactionflowmodel,yuan2026ftp}.
However, collecting real-world demonstrations for functional dexterity remains costly and noisy because each trajectory must jointly specify arm motion, multi-finger contact, and contact timing, which are difficult for an operator to perform at the same time, especially considering the embodiment gap between human and robotic hands~\cite{handaDexPilotVisionBased2019, qinAnyTeleopGeneralVisionBased2024}. Furthermore, these demonstrations usually focus on successful cases, providing limited evidence for recovery after failures.
Recovery from execution errors can be improved by real-world RL from online physical outcomes~\cite{levine2016end,luo2024serl,luo2025precise}, but direct exploration and human correction in the full arm-hand action space are inefficient for learning reactive recovery.

To address these limitations, we propose \methodname{}, a hierarchical real-world RL framework for functional dexterous manipulation.
For fine-grained finger control and precise contact execution, \methodname{} adapts kinesthetic hand-motion priors with residual RL and real contact feedback, producing functional hand skills grounded in physical interaction.
To improve exploration efficiency in high-DoF arm-hand coordination, \methodname{} freezes the hand skills as action primitives and trains a coordination policy over arm motion and skill activation, with DMP-augmented rollouts providing initial coverage.
During online fine-tuning, human corrections are provided through the hand-skill interface to make interventions more consistent and effective, while a multi-head critic improves temporal coordination and intervention advantage weighting (IAW) prioritizes useful corrections for reactive recovery.
Together, this factorization reduces the burden of full arm-hand demonstrations, exploration, and correction, enabling reliable functional dexterous manipulation with efficient online training.
%
We evaluate \methodname{} on six functional dexterous tasks across two dexterous hands, as shown in Fig.~\ref{fig:teaser}.
\methodname{} achieves 98.3\% average success within one hour of online training, outperforms IL and flat human-in-the-loop RL baselines, and remains robust to disturbances, while ablations verify the contributions of the hand-skill and coordination components.

In summary, our contributions are: (1) a hierarchical real-world RL framework that decouples high-DoF arm-hand control into functional hand-skill learning and skill-conditioned coordination, enabling efficient real-world learning of functional dexterous tasks; (2) residual-RL hand-skill learning that adapts kinesthetic priors with real contact feedback for fine-grained contact execution;
(3) skill-conditioned coordination that uses hand skills as action primitives, DMP-augmented rollouts for efficient initial coverage, a multi-head critic for temporal coordination of arm motion and hand activation, and IAW for sample-efficient online recovery learning.
\vspace{-1mm}
\section{Related Work}
\vspace{-2mm}
\noindent\textbf{Dexterous Manipulation.}
Model-based control plans stable dexterous grasps from explicit object and contact models~\cite{Miller2004graspit, Roa2015quality}, but functional manipulation depends on contact-rich dynamics that are difficult to model accurately.
Large-scale simulation has enabled robust dexterous grasping~\cite{zhang2025robustdexgrasp, chen2025clutterdexgrasp, huang2025fungrasp, lum2024dextrahg, singh2024dextrah, zhang2026congraspxl} and in-hand rotation~\cite{openai2019learningdexterousinhandmanipulation, Chen2023VisualDexterity, qi2023hand, liu2025dexndm, zhang2026unicross}, yet functional contact dynamics involving deformation, friction variation, and fine-grained force feedback remain difficult to simulate~\cite{wang2024lessons,hsieh2025learning}, causing precise contact behaviors to transfer unreliably.
IL and pretrained visuomotor policies can represent high-DoF arm-hand behaviors from demonstrations via diffusion-based policies~\cite{chi2025diffusion, Ze2024DP3}, transformer policies with action chunking~\cite{zhaoLearningFineGrainedBimanual2023,buamanee2024biactbilateralcontrolbasedimitation}, and vision-language-action models~\cite{brohan2023rt2visionlanguageactionmodelstransfer,black2026pi0visionlanguageactionflowmodel,kim2024openvlaopensourcevisionlanguageactionmodel}.
For dexterous arm-hand systems, however, such demonstrations require specialized teleoperation or capture interfaces~\cite{handaDexPilotVisionBased2019,qinAnyTeleopGeneralVisionBased2024,wangDexCapScalablePortable2024,xuDexUMIUsingHuman2025,fang2025dexop}, and human-to-robot retargeting can introduce contact noise under morphology mismatch~\cite{yinGeometricRetargetingPrincipled2025}.
Thus, demonstrations for functional dexterity remain costly and noisy, struggle to cover fine-grained finger contacts and dynamic arm-hand coordination, and provide limited recovery evidence.
\methodname{} instead obtains hand-motion priors through kinesthetic teaching~\cite{zhang2025kinedex}, adapts them into contact-aware skills with residual RL, and uses these skills to collect downstream coordination data, reducing direct teleoperation of all dexterous-hand DoFs.

\noindent\textbf{Real-World Reinforcement Learning.}
Real-world RL learns directly from physical interactions, capturing hardware dynamics and contact physics that are difficult to simulate~\cite{levine2016end,luo2021robust,ball2023efficient}.
Human-in-the-loop (HIL) methods improve sample efficiency by incorporating online corrections during training~\cite{luo2025precise,kelly2019hg,rossReductionImitationLearning2011a}, and HIL-SERL achieves near-perfect success on precise gripper tasks with demonstrations and real-time guidance~\cite{luo2025precise}.
For functional dexterous manipulation, however, the high-dimensional arm-hand system degrades data precision and efficiency while increasing exploration burden, and coupled finger joints make consistent real-time correction across all DoFs difficult~\cite{zhu2020ingredients}.
Hierarchical and skill-based methods reduce exploration complexity by reusing lower-level behaviors or guiding policy search.
Prior work chains dexterous sub-policies for long-horizon manipulation~\cite{chen2023sequential}, reuses learned rotation skills for harder in-hand reorientation~\cite{qi2025simple}, and guides RL exploration with simple sub-skill controllers~\cite{khandate2024dexterous}.
We adopt a hierarchical framework that decouples hand primitive skills from arm-hand coordination, shifting online learning from full arm-hand exploration to skill-conditioned coordination and recovery.

\section{Method}
\label{sec:method}
\vspace{-2mm}
\begin{figure}[H]
    \centering
    \includegraphics[width=\linewidth]{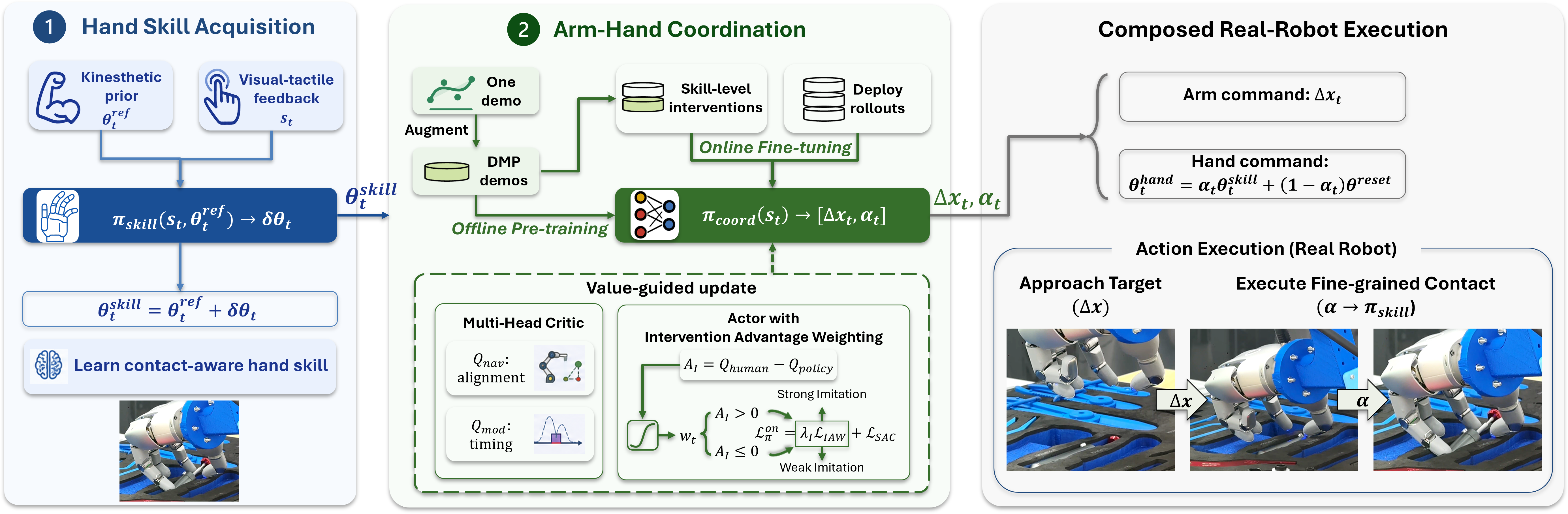}
    \caption{\textbf{Overview of \methodname{}, a hierarchical framework.}
    It learns (1) a residual-RL hand skill for contact-aware execution and (2) a coordination policy for arm motion and skill activation, initialized by DMP rollouts and refined by value-guided fine-tuning.
    At execution time, it combines arm motion and skill activation with the frozen hand skill for target approach and fine-grained contact.}
    \label{fig:pipeline}
    \vspace{-4mm}
\end{figure}


To efficiently learn reliable contact-rich dexterous manipulation policies for high-DoF arm-hand systems, \methodname{} uses a hierarchical real-world RL framework (Fig.~\ref{fig:pipeline}).
It learns (1) a residual-RL hand skill for contact-aware execution and (2) a coordination policy for arm motion and skill activation.
Coordination uses DMP rollouts for warm start, a multi-head critic to separate arm-alignment and skill-activation Q-values, and IAW for advantage-weighted intervention imitation.
We introduce formulation (Sec.~\ref{subsec:preliminaries}), hand skill training (Sec.~\ref{subsec:hand_skill}), and coordination training (Sec.~\ref{subsec:coordination}).

\subsection{Problem Formulation}
\label{subsec:preliminaries}
We formulate dexterous arm-hand manipulation as Markov Decision Processes with observation $s_t=(I_t,q_t,f_t)$, where $I_t$ contains wrist and external RGB images, $q_t$ contains arm-hand proprioception, and $f_t$ contains fingertip contact forces.
The hand skill policy $\pi_{\text{skill}}(s_t,\theta_t^{\text{ref}})$ takes a reference hand pose from a kinesthetic demonstration and outputs a residual correction $\delta\theta_t\in\mathbb{R}^{12}$:
\begin{equation}
    \theta_t^{\text{skill}} = \theta_t^{\text{ref}} + \lambda_{\text{res}}\delta\theta_t,
    \label{eq:residual_skill}
\end{equation}
where $\lambda_{\text{res}}$ limits residual exploration around the demonstrated motion.
\rebuttalrev{Here, $\theta_t^{\text{ref}}$ denotes the reference hand pose selected at execution step $t$, allowing an earlier reference state to be revisited during contact recovery (Sec.~\ref{subsec:hand_skill}).}
The coordination policy $\pi_{\text{coord}}(s_t)$ outputs $a_t^{\text{coord}}=[\Delta x_t,\alpha_t]$, where $\Delta x_t\in\mathbb{R}^6$ is the arm end-effector increment and $\alpha_t\in[0,1]$ modulates the hand skill from initial neutral hand configuration $\theta^{\text{reset}}$:
\begin{equation}
    \theta_t^{\text{hand}} = \alpha_t\theta_t^{\text{skill}} + (1-\alpha_t)\theta^{\text{reset}}.
    \label{eq:gated_hand}
\end{equation}
The robot executes $(\Delta x_t,\theta_t^{\text{hand}})$, so $\pi_{\text{skill}}$ determines how the fingers form contact and $\pi_{\text{coord}}$ determines where and when that contact is used.

\subsection{Contact-Aware Hand Skill Learning}
\label{subsec:hand_skill}
The hand-skill policy abstracts high-dimensional finger articulation into robust, contact-aware functional skills.
\rebuttalrev{We acquire a kinesthetic prior using the hand-over-hand teaching interface of~\cite{zhang2025kinedex}, recording reference hand poses $\xi_{\text{ref}}=\{\theta_0^{\text{ref}},\ldots,\theta_T^{\text{ref}}\}$ together with fingertip-force evolution and collecting per-finger force thresholds $F_{\max}^{(i)}$ for reward clipping.}
\rebuttalrev{To recover from disrupted contact, the reference can rewind to the most recent contact-consistent state when measured contact deviates substantially from the expected state.}
\rebuttalrev{Given the current observation $s_t$ and the selected reference pose $\theta_t^{\text{ref}}$, the residual policy outputs $\delta\theta_t$ and executes Eq.~\eqref{eq:residual_skill}, adjusting finger configuration and contact force to re-establish the desired contact.}
We train the critic on the executed hand command $\theta_t^{\text{skill}}$ so value estimates correspond to the actual finger configuration.

The hand skill reward combines trajectory tracking, contact incentive, and sparse skill success:
\begin{equation}
\begin{split}
    r_{\text{skill}} = &-\omega_{\text{track}}\|\theta_t^{\text{cur}}-\theta_t^{\text{ref}}\|^2
    + \omega_{\text{force}}\mathcal{F}(f_t)
    + \mathbb{I}_{\text{skill-success}}R_{\text{skill}},
\end{split}
\label{eq:skill_reward}
\end{equation}
where $\mathcal{F}(f_t)=\sum_{i=1}^{N_f}\min(\|f_t^{(i)}\|,F_{\max}^{(i)})$ sums clipped fingertip-force magnitudes.
The tracking term keeps exploration near contact geometry, the force term encourages task-relevant contact, and the sparse bonus marks operator-labeled success.
We train $\pi_{\text{skill}}$ using SAC~\cite{haarnoja2018soft} on the real robot and initialize replay with kinesthetic demonstrations.
Once converged, $\pi_{\text{skill}}$ is frozen as (i) the skill primitive invoked by $\pi_{\text{coord}}$ and (ii) a reliable intervention interface for human operators.

\subsection{Skill-Conditioned Arm-Hand Coordination}
\label{subsec:coordination}
Given the frozen hand skill, $\pi_{\text{coord}}$ 
maps $s_t$ to $[\Delta x_t,\alpha_t]$, determining \textit{where to move} the arm and \textit{when to activate} the hand skill.
We use RLPD~\cite{ball2023efficient} for sample-efficient learning from offline initialization data and online real-world interaction.
Task completion is detected by a binary classifier over ResNet-10 visual features, fingertip forces, and joint torques, producing the terminal reward $R_{\text{task}}\in\{0,1\}$.

Since the sparse reward $R_{\text{task}}$ guides approach but provides no feedback on activation timing, the modulation reward $r_{\text{mod}}$ penalizes premature activation and rewards successful activation:
\begin{equation}
r_{\text{mod}} =
\begin{cases}
    r_{\text{bonus}} & \text{if } \alpha_t>\tau \text{ and } R_{\text{task}}=1 \\
    r_{\text{penalty}} & \text{if } \alpha_t>\tau \text{ and } R_{\text{task}}=0 \\
    0 & \text{otherwise},
\end{cases}
\quad (r_{\text{penalty}}<0<r_{\text{bonus}})
\label{eq:mod_reward}
\end{equation}
\noindent\textbf{Multi-head critic.} Naively training one Q-function on $R_{\text{task}}+r_{\text{mod}}$ can make the policy avoid activation: early high-$\alpha$ trials often fail, so accumulated penalties dominate the values of high-$\alpha$ actions.
%
%
To decouple spatial approach from activation timing, we use two value heads trained with separate TD targets: a navigation head $Q_\phi^{\text{nav}}$ whose target $y^{\text{nav}}$ is built from $R_{\text{task}}$, and a modulation head $Q_\phi^{\text{mod}}$ whose target $y^{\text{mod}}$ is built from $r_{\text{mod}}$:
\begin{equation}
    \mathcal{L}_Q(\phi)=\mathbb{E}\!\left[(Q_\phi^{\text{nav}}-y^{\text{nav}})^2+(Q_\phi^{\text{mod}}-y^{\text{mod}})^2\right].
\end{equation}
Then the actor is trained with the combined value: $\hat{Q}_\phi(s,a)=Q_\phi^{\text{nav}}(s,a)+Q_\phi^{\text{mod}}(s,a).$
During actor optimization, we evaluate \(Q^{\mathrm{nav}}\) with gradients stopped through \(\alpha_t\), and evaluate \(Q^{\mathrm{mod}}\) with gradients stopped through \(\Delta x_t\). This encourages the navigation head to shape arm alignment and the modulation head to shape activation timing, while both heads score the same physical action. With this multi-head critic, coordination training follows an offline-to-online scheme.

\noindent\textbf{Offline pre-training.}
%
To avoid the prohibitive sample complexity of learning from scratch, we bootstrap coordination from one demonstration containing an arm trajectory and skill-activation schedule.
Cartesian Dynamic Movement Primitives (DMP)~\cite{ude2014orientation} generate diverse arm trajectories by randomizing initial poses while fixing the goal.
DMP action labels use clipped waypoint differences for $\Delta x_t$ and align $\alpha_t$ with the demonstration by task progress.
The resulting $\mathcal{D}_{\text{dmp}}$ provides initial-state coverage.
We pre-train $\pi_{\text{coord}}$ with SAC plus a global behavior-cloning term:
\begin{equation}
\begin{aligned}
    \mathcal{L}^{\text{off}}_{\pi}
    =&\ \mathbb{E}_{s\sim\mathcal{D}_{\text{dmp}},a\sim\pi}
    [\eta\log\pi(a|s)-\hat{Q}_\phi(s,a)] \\
    &+\lambda_{\text{gbc}}\mathbb{E}_{(s,a^{\text{dmp}})\sim\mathcal{D}_{\text{dmp}}}
    [\|\bar{a}_{\pi}(s)-a^{\text{dmp}}\|^2],
\end{aligned}
\label{eq:offline_actor}
\end{equation}
where $a^{\text{dmp}}$ and $\bar{a}_{\pi}$ denote the DMP label and deterministic policy action.
This keeps the policy near generated approach-and-activation behavior while allowing critic improvement.

\noindent\textbf{Online fine-tuning.}
Online training uses real policy rollouts and skill-level corrections to refine $\pi_{\text{coord}}$ for better target alignment, skill timing, and recovery under real contact dynamics.
Arm interventions use relative SE(3) retargeting through Gello~\cite{wu2024gello} for smooth takeover, while hand interventions trigger the frozen $\pi_{\text{skill}}$ by button press rather than teleoperating all hand DoFs.
Rollouts and intervention segments are stored for RLPD updates, and IAW weights an intervention-imitation loss by the critic-estimated advantage of each correction.
For an intervention action $a^I$, we estimate its advantage over the current deterministic action using the target critic ensemble:
\begin{equation}
    A^I(s)=\min_i \hat{Q}_{\bar{\phi},i}(s,a^I)-\min_i \hat{Q}_{\bar{\phi},i}(s,\bar{a}_{\pi}(s)).
    \label{eq:intervention_advantage}
\end{equation}
The imitation weight and loss are
\begin{equation}
    w(s)=\sigma(\mathrm{Norm}(A^I(s))/\tau_I),\qquad
    \mathcal{L}_{\text{IAW}}=\mathbb{E}_{(s,a^I)\sim\mathcal{D}_I}
    [w(s)\|\bar{a}_{\pi}(s)-a^I\|^2].
    \label{eq:iaw_loss}
\end{equation}
The online actor objective augments the standard SAC actor loss with the IAW term,
$\mathcal{L}^{\text{on}}_{\pi}=\mathcal{L}_{\text{SAC}}+\lambda_I\mathcal{L}_{\text{IAW}}$.
IAW prevents blind cloning: corrections estimated to improve over the current policy receive larger weights, while low-advantage corrections have limited influence.

\section{Experiments}
\label{sec:experiments}

\subsection{Experimental Setup}
\label{sec:tasks}
\label{sec:exp_details}
We evaluate \methodname{} on six real-world tasks spanning Power-Drill Actuation, Angle-Spreader Retrieval, Thin-Handle Retrieval, Power-Drill Bit Removal, Pipette Dispensing, and Thin-Shaft Tool Retrieval, corresponding to \hyperref[fig:teaser]{Fig.~\ref*{fig:teaser}A--F} in order.
These tasks cover precise tool-use, contact-rich rotation, and constrained-space manipulation.
A trial is counted as successful only when the task-specific functional outcome is completed: tightening the target screw (A), fully extracting the target tools (B,C,F), removing the drill bit (D), or dispensing into the vial (E).
All tasks use Franka Research 3 arms~\cite{FRANKA-RESEARCH-3} with either a Sharpa hand~\cite{Sharpa} or an XHand~\cite{Xhand}, wrist RGB observations from an Intel RealSense D405, and a side-view D415 when needed.
\begin{table*}[t]
\centering
\resizebox{\textwidth}{!}{%
\begin{tabular}{l|c|c|c|ccc|cc}
\toprule
\multirow{2}{*}{\textbf{Task}}
& \multirow{2}{*}{\shortstack{\rule{0pt}{8pt}\textbf{IL Data}\\ \textbf{Collection}\\ \textbf{(mins)}}}
& \multirow{2}{*}{\shortstack{\rule{0pt}{8pt}\textbf{RL Data}\\ \textbf{Collection}\\ \textbf{(mins)}}}
& \multirow{2}{*}{\shortstack{\rule{0pt}{8pt}\textbf{RL Online}\\ \textbf{Training}\\ \textbf{(mins)}}}
& \multicolumn{5}{c}{\textbf{Success Rate (\%)}} \\
\cline{5-9}
& & &
& \shortstack{\rule{0pt}{8pt}HIL-SERL\\ (Glove+Gello)}
& \shortstack{\rule{0pt}{8pt}HIL-SERL\\ (Ours Interv.)}
& \shortstack{\rule{0pt}{8pt}\methodname\\ (Ours)}
& \shortstack{\rule{0pt}{8pt}Behavior\\ Cloning (BC)}
& \shortstack{\rule{0pt}{8pt}Diffusion\\ Policy (DP)} \\
\midrule
\multicolumn{9}{c}{\textbf{Sharpa Main Validation Tasks}} \\ \midrule
Power-Drill Actuation & 150 & 34 & 60 & 15 & 62\,|\,30$^\ddagger$ & \textbf{100} & 15 & 30 \\
Angle-Spreader Retrieval & 60 & 13 & 55 & -- & 40\,|\,0$^\ddagger$ & \textbf{100} & 30 & 60 \\
Thin-Handle Retrieval & 80 & 17 & 60 & -- & 28\,|\,0$^\ddagger$ & \textbf{100} & 35 & 40 \\
Power-Drill Bit Removal & 80 & 21 & 22 & -- & 60\,|\,20$^\ddagger$ & \textbf{100} & 30 & 36 \\ \midrule
\multicolumn{9}{c}{\textbf{XHand Portability Tasks}} \\ \midrule
Pipette Dispensing & 100 & 14 & 60 & 12 & 60\,|\,32$^\ddagger$ & \textbf{90} & 12 & 26 \\
Thin-Shaft Tool Retrieval & 100 & 14 & 20 & -- & 30\,|\,0$^\ddagger$ & \textbf{100} & 20 & 58 \\
\midrule
Average & 95.0 & 18.8 & 46.2 & 4.5 & 13.7 & \textbf{98.3} & 23.7 & 41.7 \\
\bottomrule
\end{tabular}%
}
\caption{\textbf{Main results.}
Success rates are computed over 50 trials. ``--'' marks HIL-SERL (Glove+Gello) settings where direct arm-hand teleoperation can not consistently provide reliable corrections for contact-rich tasks. $^\ddagger$a\,|\,b reports arm approach success ($a$) and final task success with the required hand contact ($b$). Averages use final-task success and count ``--'' as 0.}
\label{tab:main_results}
\end{table*}

\noindent\textbf{Metrics.}
For final evaluation, each task randomizes object poses within the training workspace while using the same robot reset strategy as training.
We report IL data-collection time for recording 200 demonstrations, RL data-collection time for generating 100 automated DMP rollouts without human labor, RL online training time for HIL fine-tuning of the coordination policy, and success rate, the percentage of evaluation trials that meet the task criterion.
Full evaluation protocol details are provided in the Appendix.

\noindent\textbf{Baselines and Protocol.}
\label{sec:baselines_protocol}
To evaluate \methodname{}, we compare against HIL-SERL~\cite{luo2025precise}, a human-in-the-loop RL method that combines offline data, online RL, and real-time interventions for real-world manipulation.
\rebuttalrev{The key difference is that \methodname{} packages contact execution into a frozen hand skill and trains a hierarchical coordination policy over arm motion and skill activation, whereas HIL-SERL keeps a flat policy architecture that directly outputs full-DoF arm-hand actions.}

\textbf{HIL-SERL (Glove+Gello)} uses AnyTeleop~\cite{qinAnyTeleopGeneralVisionBased2024} for hand-mapped teleoperation and Gello~\cite{wu2024gello} for arm control, testing the benefit of the hand-skill intervention interface over full arm-hand correction.
\textbf{HIL-SERL (Ours Interv.)} uses the same learned intervention interface as \methodname{} but keeps a flat full-DoF policy, isolating the benefit of the hierarchical architecture beyond the intervention mechanism.
\rebuttalrev{All RL methods in Tab.~\ref{tab:main_results} are initialized from an offline demo buffer containing 100 DMP-generated trajectories and trained online for up to one hour, a cap that pilot runs found sufficient while limiting variability from operator attention and intervention timing.}
\rebuttalrev{Three operators trained and evaluated the six tasks. Final success rates differed by approximately 2--3 percentage points across operators, although their intervention styles produced modest behavioral variation, suggesting that performance does not depend strongly on a single operator.}

\rebuttalrev{We additionally evaluate \textbf{Skill-HIL}, which equips standard HIL-SERL with the same frozen hand skill for robot execution, to test whether skill reuse alone explains \methodname{}'s gains.}
\rebuttalrev{Although the skill reduces the burden of finger-level exploration and correction, coordination learning still requires credit assignment between arm alignment and skill activation, as well as effective learning from heterogeneous human interventions.}
\rebuttalrev{The controlled comparison in Appendix~\ref{appendix:skill_hil} shows that sharing the hand skill alone does not recover \methodname{}'s performance under matched task settings.}

For IL baselines, \textbf{Behavior Cloning (BC)}~\cite{pomerleau1988alvinn} and \textbf{Diffusion Policy (DP)}~\cite{chi2025diffusion} are trained with 200 successful teleoperated demonstrations per task to test whether offline imitation from human demonstrations can handle functional dexterous tasks, where small contact or timing deviations often cause failure.
To distinguish training-data coverage from the learning paradigm, the Appendix reports additional IL baselines trained on the same successful trajectory sources as \methodname{}.
\vspace{-2mm}
\subsection{Main Results}
\label{sec:main_results}
\vspace{-1mm}
We evaluate the performance of \methodname by comparing it against representative baselines that include both RL and IL paradigms.
Tab.~\ref{tab:main_results} shows the strongest performance: 98.3\% average success across six tasks and two hands with 18.8 minutes of RL data collection and 46.2 minutes of online training.
To further validate the autonomy of \methodname, we monitor the intervention rate over time, defined as the ratio of intervention steps to total steps per episode.
In the Appendix, intervention-rate curves show human guidance approaching zero by the end of training, indicating autonomous execution.

Despite sharing the same RL backbone and initial data, \methodname{} substantially outperforms the HIL-SERL baselines, which we attribute to intervention quality and exploration efficiency.
Specifically, standard teleoperation methods (Glove+Gello) generate noisy interventions due to morphology mismatches and retargeting errors, leading to inconsistent demonstrations (for instance, wrist shaking when operating a power drill) and a degraded replay buffer.
For tool retrieval tasks with complex hand motions, obtaining successful corrections becomes exceedingly challenging, leading to complete failures marked ``--'' in Tab.~\ref{tab:main_results}.
Using our hand-skill intervention interface improves correction quality, as HIL-SERL (Ours Interv.) rises to 13.7\% average success, but its flat full-DoF policy remains far below \methodname{}.
The staged success results marked with $\ddagger$, such as 40\,|\,0, 28\,|\,0, and 30\,|\,0, show that this flat policy often approaches the target but fails to complete reliable hand contact.
This indicates that the absence of hierarchical decomposition still limits exploration efficiency, particularly for tasks with complex hand motions.
%
%
For example, a flat policy for \textit{Thin-Handle Retrieval} must discover sustained non-prehensile levering, precision grasping, and collision avoidance as one multi-stage contact sequence,
which requires many full arm-hand interactions, producing many noisy rollouts before the correct contact sequence is found.
Overall, the results indicate that our hierarchical structure effectively improves intervention quality by leveraging fine-grained hand skill primitives and enables more efficient learning by simplifying the exploration space. 

Furthermore, even with 200 expert demonstrations per task, IL baselines struggle with precise and contact-rich tasks. BC suffers from compounding error in long-horizon execution, reaching only 15\% on \textit{Power-Drill Actuation} and 12\% on \textit{Pipette Dispensing}. DP improves by modeling multimodal action distributions, achieving 60\% on \textit{Angle-Spreader} and 58\% on \textit{Thin-Shaft Tool}, but still fails on the \textit{Bit Removal} (36\%) and \textit{Pipette} (26\%) tasks. These failures highlight deficiencies in fine-grained alignment and reactive behavior, indicating that successful offline demonstrations weakly cover contact deviations and provide limited recovery signal.
To separate data coverage from the learning paradigm, we additionally report IL baselines in the Appendix trained with successful trajectories from \methodname{}.
These policies remain brittle under precise alignment requirements and off-nominal contact states, indicating that purely offline training makes them sensitive to distribution shifts during inference. Our method addresses this issue by incorporating online exploration and human interventions to refine closed-loop behaviors. 
In summary, the results show that our method achieves both more efficient and effective learning of contact-rich functional dexterous manipulation tasks compared with the baselines.

\begin{figure}[!t]
    \centering
    \includegraphics[width=\linewidth]{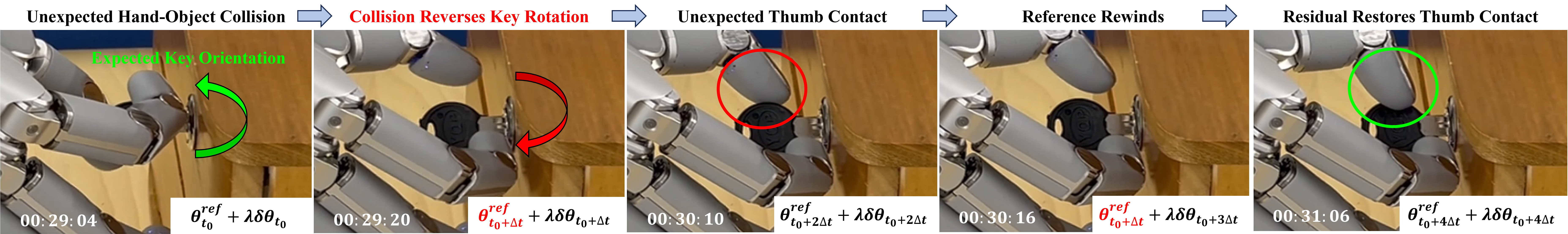}
    \ifshowrebuttalchanges
      \captionsetup{labelfont={color=red},textfont={color=red}}
    \fi
    \caption{\textbf{Finger-level contact recovery.} Following collision-induced key reorientation, the hand restores thumb contact without arm re-alignment. Red and green circles mark disturbed and restored contact states, respectively.}
    \label{fig:finger_recovery}
\end{figure}

\noindent\textbf{Robustness Evaluation.}
\rebuttalrev{We evaluate recovery from task-level perturbations and finger-level contact disturbances.}
\rebuttalrev{For the full system, we conduct 20 perturbation trials per task, covering randomized object poses across all tasks, approach-phase displacements in alignment-sensitive tasks, and drill-chuck re-tightening during \textit{Power-Drill Bit Removal}.}
\rebuttalrev{\methodname{} succeeds in 20/20 trials on four of the six tasks and 18/20 on both \textit{Power-Drill Actuation} and \textit{Pipette Dispensing}.}
\rebuttalrev{The remaining failures involve screw--drill angular misalignment or vial motion after alignment.}
\rebuttalrev{Successful trials show that \methodname{} re-aligns before contact and suspends hand-skill activation until contact becomes reachable (Fig.~\ref{fig:teaser}), consistent with recovery through adaptive activation timing rather than fixed replay.}
\rebuttalrev{Additional recovery examples are provided in the videos on the \href{https://hly-123.github.io/HiFun/}{project website}.}

\rebuttalrev{To isolate finger-level adaptation, we additionally fix the arm near the execution region and evaluate the learned hand skills under unseen object orientations and external force disturbances.}
\rebuttalrev{Under orientation changes, \textit{Angle-Spreader Retrieval} achieves 19/20 success with toolbox yaw of $0$--$30^{\circ}$ and tool pitch/roll of $0$--$20^{\circ}$, while \textit{Power-Drill Bit Removal} achieves 18/20 with drill yaw up to $30^{\circ}$.}
\rebuttalrev{Under external force disturbances, the fixed-arm \textit{Angle-Spreader Retrieval} skill achieves 18/20 success.}
\rebuttalrev{Fig.~\ref{fig:finger_recovery} illustrates this recovery mechanism: after an unexpected collision reorients the key and disrupts thumb contact, reference-state rewind and residual finger corrections restore contact without arm re-alignment.}
\rebuttalrev{The supplementary video also shows the same hand skill operating on 6.5\,cm and 5.5\,cm tool variants after learning with a 7.5\,cm tool.}
\rebuttalrev{Together, these results support local finger-level contact recovery and transfer across orientation and size variations around the demonstrated contact structure.}

\vspace{-1mm}
\subsection{Ablation and Analysis}
\label{sec:Ablation}
\vspace{-1mm}
\noindent\textbf{1) Arm-Hand Coordination:}
Tab.~\ref{tab:high_level_analysis} analyzes coordination policy training from two aspects: (i) the effect of warm-start data sources (DMP rollouts vs. human demonstrations) and (ii) the contributions of the multi-head critic and intervention advantage weighting (IAW) through component ablations.
%
\begin{table}[t]
\centering
\setlength{\tabcolsep}{3pt}
\resizebox{0.73\linewidth}{!}{%
\begin{tabular}{l|cc|cc}
\toprule
\multicolumn{5}{c}{\textbf{Data Source Comparison}} \\ \midrule
\multirow{2}{*}{\textbf{Task}} & \multicolumn{2}{c}{\textbf{Success Rate (\%)}} & \multicolumn{2}{c}{\textbf{Data Collection Time (mins)}} \\
\cmidrule{2-5}
 & \textbf{(w/ DMP)} & \textbf{(w/ Human)} & \textbf{(w/ DMP)} & \textbf{(w/ Human)} \\ \midrule
Angle-Spreader Retrieval & 100 & 85 & 13 & 60 \\
Power-Drill Bit Removal & 100 & 100 & 21 & 80 \\ \midrule
\multicolumn{5}{c}{\textbf{Component Ablation (Success Rate \%)}} \\ \midrule
\textbf{Task} & \textbf{Ours (Full)} & \textbf{w/o Multi-head} & \textbf{w/o IAW (Uniform BC)} & \textbf{w/o Interv. BC} \\ \midrule
Angle-Spreader Retrieval & 100 & 85 & 60 & 40 \\
Power-Drill Bit Removal & 100 & 75 & 60 & 40 \\ \bottomrule
\end{tabular}%
}
\vspace{2mm}
\caption{\textbf{Coordination policy analysis.}
Top: DMP rollouts vs. human demonstrations for warm-start data. Bottom: ablations of the multi-head critic and intervention advantage weighting (IAW).}
\label{tab:high_level_analysis}
\end{table}

\noindent\textbf{Data Source Analysis.}
%
Compared with human demonstrations, DMP rollouts achieve the same or higher final success (100\%) with about $4\times$ less collection time (Tab.~\ref{tab:high_level_analysis}), indicating that automatically collected DMP data can serve as a practical alternative to labor-intensive human demonstrations.
Additional trajectory-distribution analysis is provided in the Appendix.

\noindent\textbf{Effect of Multi-head Critic.}
Tab.~\ref{tab:high_level_analysis} shows that the multi-head critic is important for complex arm-skill coordination: removing it drops success from 100\% to 85\% on \textit{Angle-Spreader} and to 75\% on \textit{Bit Removal}.
To explain this drop, we measure how close the arm end-effector is to the target when the trained policy triggers the hand skill.
The multi-head critic keeps 95\% of activations within the $2.00\,\mathrm{cm}$ pre-contact region, while the single-head variant increases the 95th-percentile error to $4.58\,\mathrm{cm}$.
This indicates that a single-head critic activates the skill before the arm is sufficiently aligned, because it conflates sparse terminal success rewards with smaller modulation penalties.
Simply increasing the penalties does not resolve the conflict, as it can over-suppress activation to avoid penalty accumulation.
By decoupling arm-alignment and skill-activation value learning, the multi-head critic enables efficient arm movement with better-timed hand-skill activation.

%
\noindent\textbf{Effect of IAW.}
During online fine-tuning, intervention actions provide an additional imitation loss for the actor objective.
IAW weights each intervention-imitation loss by the critic-estimated advantage in Eq.~\eqref{eq:intervention_advantage}, comparing the human correction with the current policy action.
We compare it with two variants: \textit{Uniform Interv. BC}, which keeps the imitation loss but assigns all interventions equal weight, and \textit{w/o Interv. BC}, which removes this imitation loss from the actor objective.
\textit{Uniform Interv. BC} drops success to 60\% and \textit{w/o Interv. BC} to 40\%, while \methodname{} attains lower cycle time (6.69\,s) than both variants (9.21\,s and 10.03\,s).
These results show that intervention imitation helps, but uniform cloning can copy noisy or lower-value corrections and pull the actor away from Q maximization.
The Appendix further shows that IAW assigns separated weights to weak and strong corrections (0.175 vs. 0.884 mean), with high-weight corrections occurring where the current policy action has lower estimated value.
Thus, IAW concentrates imitation on states where the policy needs guidance while limiting low-advantage corrections that could conflict with the critic objective.

\begin{table}[t]
\centering
\setlength{\tabcolsep}{3pt}
\resizebox{0.7\linewidth}{!}{%
\begin{tabular}{l|c|c|c|c}
\toprule
\textbf{Task} & \textbf{Ours ($\pi_{skill}$)} & \textbf{Kinesthetic Replay} & \textbf{DexPilot$^{*}$} & \textbf{AnyTeleop$^{*}$} \\ \midrule
Power-Drill Actuation & 20/20 & 0/20 & 18/20 & 19/20 \\
Angle-Spreader Retrieval & 20/20 & 0/20 & 7/20 & 5/20 \\
Thin-Handle Retrieval & 20/20 & 0/20 & 0/20 & 0/20 \\
Power-Drill Bit Removal & 20/20 & 5/20 & 3/20 & 3/20 \\ \midrule
Pipette Dispensing & 20/20 & 0/20 & 20/20 & 20/20 \\
Thin-Shaft Tool Retrieval & 20/20 & 0/20 & 0/20 & 0/20 \\ \bottomrule
\end{tabular}%
}
\vspace{2mm}
\caption{\textbf{Hand Motion Control Comparison.} Success rate over 20 trials.}
\label{tab:low_level}
\end{table}
\noindent\textbf{2) Hand Motion Control Comparison:}
To isolate contact execution, Tab.~\ref{tab:low_level} compares the learned residual hand skill $\pi_{\text{skill}}$ with open-loop kinesthetic replay and retargeted teleoperation, with DexPilot$^{*}$ and AnyTeleop$^{*}$ reported after task-specific tuning.
Ours succeeds in all 20 trials on all tasks, while Kinesthetic Replay succeeds only partially on Power-Drill Bit Removal (5/20) and fails elsewhere because geometric trajectories without force feedback cannot sustain precision contact or constrained extraction; retargeted teleoperation reaches 20/20 on Pipette but drops on constrained retrieval and drill-bit removal due to fine-grained thumb-finger retargeting errors.
Thus, residual contact modulation is necessary for reliable functional hand skills across both embodiments.

%
\vspace{-2mm}
\section{Conclusion}
\vspace{-2mm}
%
Functional dexterous manipulation on real robots is difficult because precise contact, arm-hand coordination, and recovery are tightly coupled in a high-dimensional control space.
We present \methodname{}, a hierarchical real-world RL framework that learns contact-aware hand skills separately and freezes them for coordination and human intervention.
The resulting skill-conditioned policy optimizes arm motion, skill activation, and recovery without directly exploring or correcting all arm-hand DoFs.
DMP-augmented rollouts warm-start coordination, the multi-head critic improves temporal arm-hand coordination, and IAW prioritizes useful interventions for recovery learning.
Across six real-world tasks and two dexterous hands, \methodname{} achieves 98.3\% average success within one hour of online training and remains robust to dynamic disturbances.
These results suggest that structuring dexterous learning around functional hand skills is a practical path toward efficient, reliable real-world manipulation with multi-fingered hands.

\vspace{-2mm}
\section{Limitations}
\vspace{-2mm}
\methodname{} assumes task-relevant hand skills learned from kinesthetic priors and real contact feedback.
This improves fine-grained contact execution and reduces full arm-hand exploration, but tasks demanding contact behaviors beyond the learned skill set may fail and require additional priors, online adaptation, or reusable hand-skill libraries.
%
Our method also targets reliable low-level execution rather than open-ended planning or language-conditioned generalization.
This scope makes it complementary to VLA policies and world models: HiFun could provide contact-rich low-level controllers and consistent autonomous rollouts, while pretrained models provide task selection and semantic grounding.
Combining pretrained reasoning with \methodname{} controllers remains future work.

\clearpage
\bibliography{references}
\clearpage
\providecommand{\appendices}{}
\appendix
\appendices

\section{Method Details}
\label{appendix:method_details}

\noindent\textbf{Trajectory augmentation via Cartesian DMP.}
\label{appendix:dmp}
To provide offline data for coordination pre-training, we generate rollouts from a single reference demonstration instead of collecting full arm-hand demonstrations for every rollout.
The reference contains an arm end-effector trajectory $\{[\mathbf{p}^{ref}_t,\mathbf{q}^{ref}_t]\}_{t=0}^{T}$ and a hand-skill activation schedule $\{\alpha^{ref}_t\}_{t=0}^{T}$ collected with a Gello interface~\cite{wu2024gello}.
We fit Cartesian Dynamic Movement Primitives (DMP)~\cite{ude2014orientation} to the demonstrated arm trajectory.
The translational dynamics follow a second-order spring-damper system with a learned forcing term $\Phi_p(s)$, while the rotational dynamics evolve on the quaternion manifold:
\begin{align}
    \lambda^2 \ddot{\mathbf{p}} &= K_p (\mathbf{g}_p - \mathbf{p}) - \lambda D_p \dot{\mathbf{p}} + \Phi_p(s), \label{eq:dmp_trans} \\
    \lambda \dot{\bm{\eta}} &= K_o \big(2 \log(\mathbf{g}_o * \bar{\mathbf{q}})\big) - D_o \bm{\eta} + \Phi_o(s), \label{eq:dmp_rot1} \\
    \lambda \dot{\mathbf{q}} &= \tfrac{1}{2}\, \bm{\eta} * \mathbf{q}. \label{eq:dmp_rot2}
\end{align}
Here $\lambda$ is the temporal scaling factor, $s$ is a canonical phase variable, $\mathbf{g}_{\{p,o\}}$ are goal configurations, and $K_{\{p,o\}},D_{\{p,o\}}$ are stiffness and damping gains.
The forcing terms $\Phi_{p,o}(s)$ encode the demonstrated motion profile.
To augment the data, we fix the goal and randomize the initial pose $[\mathbf{p}_0,\mathbf{q}_0]$ within the task workspace before integrating Eq.~\eqref{eq:dmp_trans}--\eqref{eq:dmp_rot2}.
Since DMP modifies only the arm motion, the hand activation follows the demonstration by phase alignment, $\alpha_t=\alpha^{ref}(s_t)$.

We execute the generated trajectories on the physical robot to collect state-action transitions.
The arm action label is the clipped displacement between the next DMP waypoint $x^*_{t+1}$ and the current robot state $x_t$:
\begin{equation}
    \Delta x_t = \mathrm{clip}\big(x^*_{t+1}-x_t,\; -1,\; 1\big).
    \label{eq:dmp_action}
\end{equation}
When $\alpha_t>\tau$, the frozen hand-skill policy $\pi_{\text{skill}}$ generates finger commands following Eq.~\eqref{eq:residual_skill}; otherwise the hand remains at $\theta^{\text{reset}}$.
The collected transitions form $\mathcal{D}_{\text{dmp}}$, which is used for offline pre-training and as seed data in the online replay buffer.

\noindent\textbf{DMP rollout coverage analysis.}
The comparison illustrates why DMP augmentation is used as the offline data source for coordination pre-training.
Human demonstrations contain realistic task executions but are time-consuming and vary with operator timing, while DMP rollouts systematically cover different approach states from the same task goal.
This gives the coordination policy a varied warm start without repeated human arm-hand demonstrations.
\begin{figure}[H]
    \centering
    \includegraphics[width=0.82\linewidth]{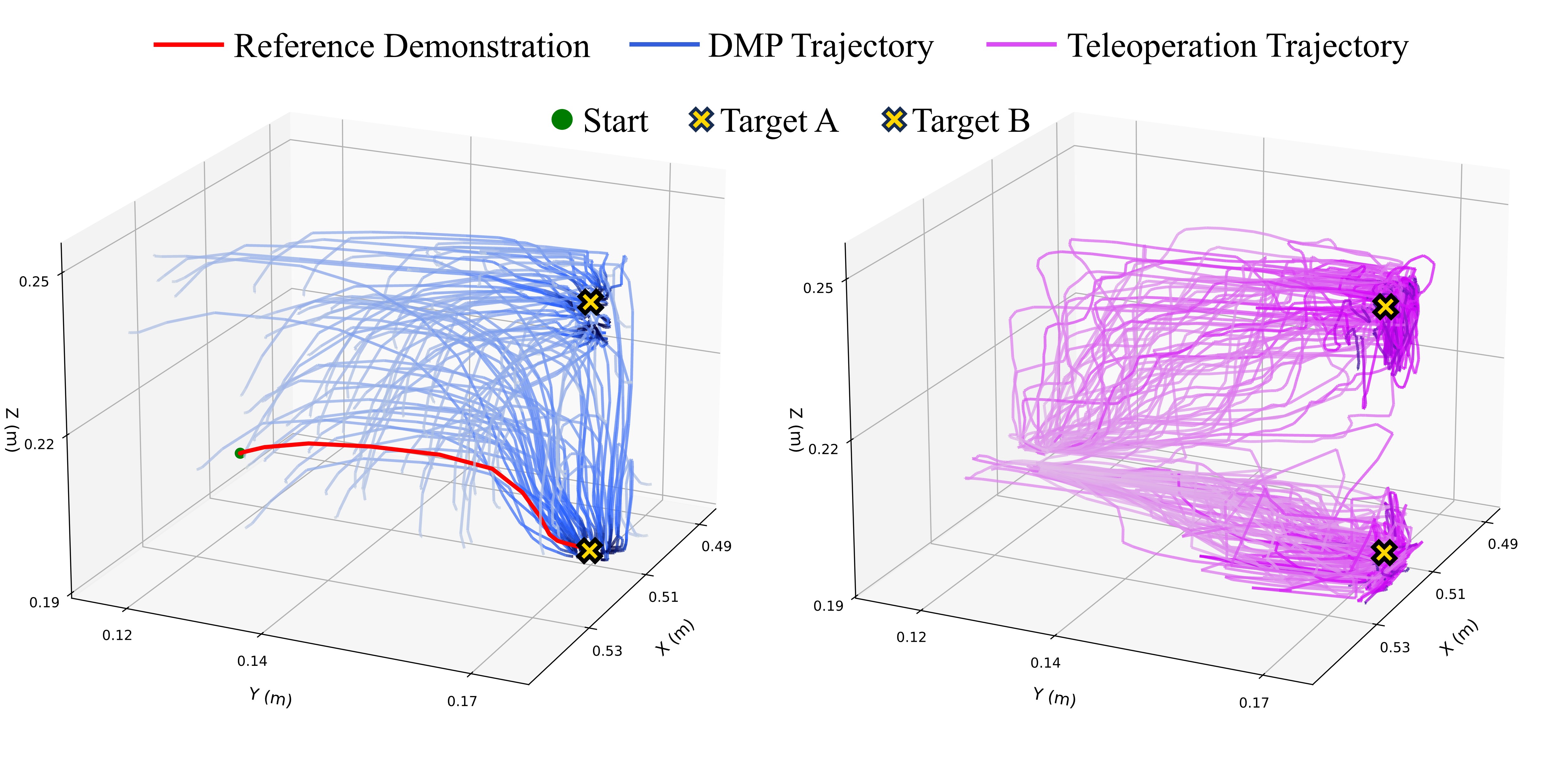}
    \caption{\textbf{Trajectory distribution from DMP rollouts and human demonstrations.}
    Starting from one reference trajectory, DMP augmentation randomizes initial arm poses while preserving the demonstrated goal and activation progress.
    This broadens approach coverage beyond repeated human demonstrations.}
    \label{fig:dmp}
\end{figure}

\noindent\textbf{SAC-based offline pre-training.}
\label{app:offline_pretrain}
We initialize the coordination policy $\pi_{\text{coord}}$ with Soft Actor-Critic (SAC)~\cite{haarnoja2018soft} rather than specialized offline RL objectives such as CQL~\cite{kumar2020conservative} or IQL~\cite{kostrikov2022offline}.
This choice follows from the structure of $\mathcal{D}_{\text{dmp}}$: unlike standard offline RL datasets with mixed-quality behavior, the DMP dataset is generated from successful reference motions and mainly provides coarse coverage for later online fine-tuning.
In this setting, we empirically find that IQL behaves similarly to behavior cloning and CQL adds little benefit when the data already covers successful state regions.

\begin{figure}[H]
    \centering
    \includegraphics[width=0.68\linewidth]{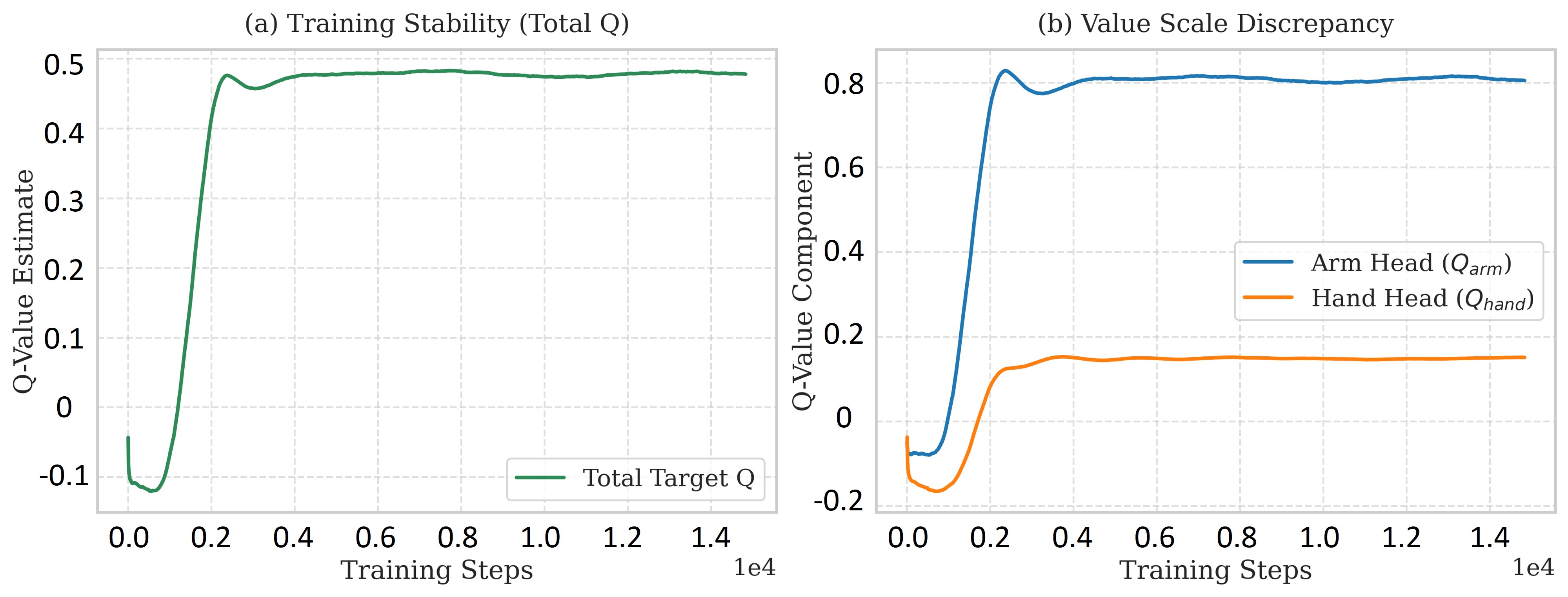}
    \caption{\textbf{Offline pre-training analysis on a representative disassembly task.}
    (a) The total target Q-value converges without value divergence under SAC.
    (b) Decomposed Q-values reveal a scale gap between arm alignment and hand activation, motivating the multi-head critic.}
    \label{fig:pretrain_curves}
\end{figure}

Figure~\ref{fig:pretrain_curves} validates this choice on a representative disassembly task.
The total target Q-value remains stable during offline pre-training, and the separated critic heads reveal a clear scale gap between arm-related and hand-activation values.
Using SAC for both offline pre-training and online fine-tuning also avoids a value-calibration gap between pessimistic offline objectives and real online returns~\cite{nakamoto2023calql,nair2020awac}.

\section{Experiment Details}
\label{appendix:eval_protocol}

\subsection{Hardware Setup}
\begin{figure}[H]
    \centering
    \includegraphics[width=\linewidth]{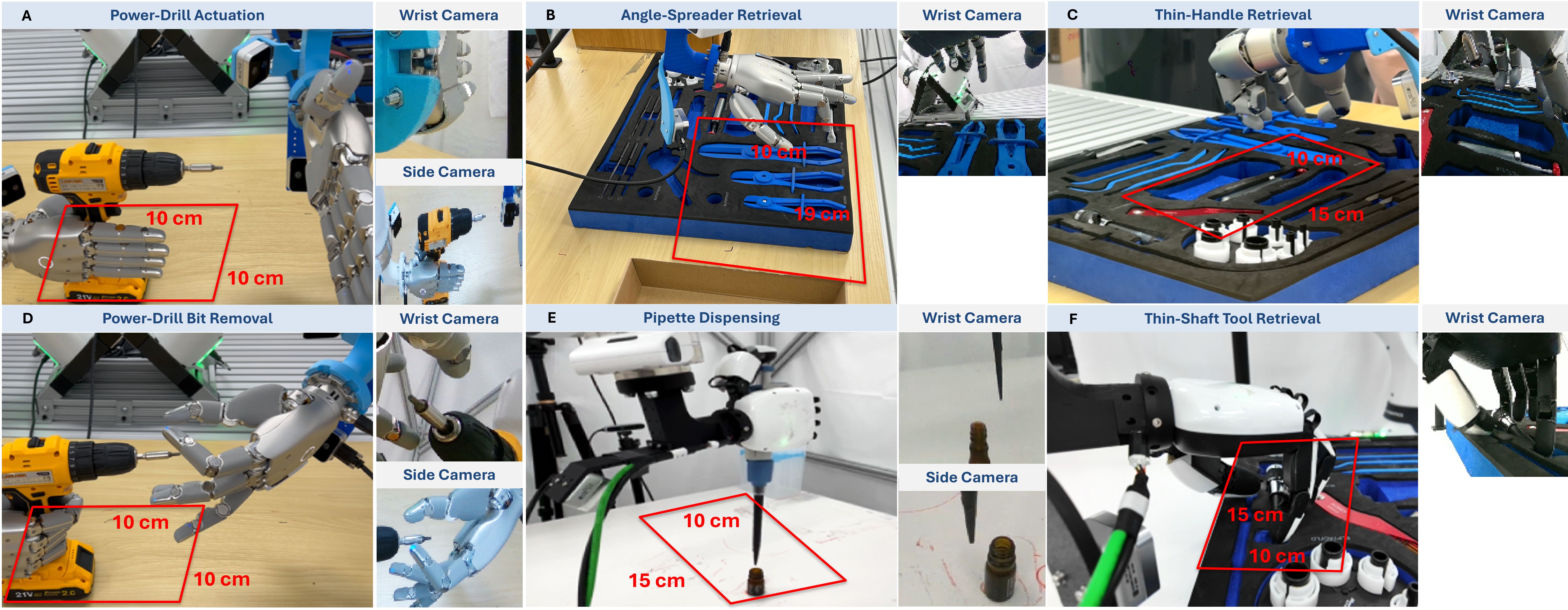}
    \caption{\textbf{Task setup and camera views.}
    The figure shows the six task setups (A--F), the workspace regions used for training and evaluation, and the corresponding wrist and side camera observations.}
    \label{fig:task}
\end{figure}

All tasks use Franka Research 3 arms with either a Sharpa hand or an XHand.
The setup includes wrist RGB observations from an Intel RealSense D405 and a side-view D415 when needed.
Figure~\ref{fig:task} summarizes the task setup and camera views across all six tasks.

\subsection{Task Descriptions}
The descriptions below define the task sequence, success condition, and main difficulty for each task, with (A)--(D) serving as Sharpa main validation tasks and (E)--(F) as XHand portability tasks.

\textbf{(A) Power-Drill Actuation} (Precise Tool-Use) requires maintaining a stable functional grasp on the drill, aligning the screwdriver bit with the target screw, and actuating the trigger to tighten the screw into the fixture.
Success requires the target screw to be tightened into the fixture.
The main difficulty is that axial alignment must be established before activation, since premature trigger engagement or excessive contact force can shift the hand pose and cause misalignment.

\textbf{(B) Angle-Spreader Retrieval} (Constrained-Space Manipulation) requires extracting the target tool from a constrained tray where surrounding slots limit direct grasping.
Success requires the target tool to be fully extracted.
The main difficulty is creating enough fingertip contact to expose or lift the tool before forming a stable grasp.

\textbf{(C) Thin-Handle Retrieval} (Constrained-Space Manipulation) requires extracting a thin-handle tool from a narrow opening where limited clearance prevents direct grasping.
Success requires the target tool to be fully extracted.
The main difficulty is executing non-prehensile levering to create access, followed by a precision grasp while avoiding collisions with the surrounding tray.

\textbf{(D) Power-Drill Bit Removal} (Contact-Rich Rotation) requires grasping the drill chuck, applying sustained rotational motion to loosen it, and removing the drill bit.
Success requires the drill bit to be removed from the chuck.
The main difficulty is maintaining stable multi-finger contact under friction and changing contact forces during rotation.

\textbf{(E) Pipette Dispensing} (Precise Tool-Use) requires aligning the pipette tip with a vial and depressing the plunger.
Success requires dispensing liquid into the vial without collision or leakage.
The main difficulty is applying thumb force while preserving grasp stability and keeping the tip aligned with the vial opening.

\textbf{(F) Thin-Shaft Tool Retrieval} (Constrained-Space Manipulation) requires extracting a thin-shaft tool embedded in foam with limited clearance.
Success requires the target tool to be fully extracted.
By using a contact sequence similar to Thin-Handle Retrieval with the XHand embodiment, this task tests cross-hand deployment for constrained-space retrieval.

\subsection{Evaluation Protocol}
\noindent\textbf{Main Evaluation.}
For main evaluation, each task is tested over 50 trials.
For every task, object poses are randomized within the red-box workspace shown in Fig.~\ref{fig:task}, and robot reset poses follow the train-time reset protocol.
This evaluates repeatable functional execution within the trained workspace rather than zero-shot workspace generalization.
Full success requires the task-specific functional outcome to be completed: tightening the target screw, fully extracting the target tool, removing the drill bit, or dispensing into the vial.
Reported success rate is the mean success credit over 50 trials: full success receives credit 1 and failure receives credit 0.
For rare baseline trials that reach the task-specific final contact stage but terminate at the evaluation horizon before completing the functional outcome, we assign credit 0.5.
This distinguishes horizon-limited near-completion from failures that never establish the required contact.

\noindent\textbf{Perturbation Evaluation.}
The robustness evaluation uses 20 perturbation trials per task.
The perturbations include randomized object poses across tasks, approach-phase displacements for alignment-sensitive tasks, and drill-chuck re-tightening for \textit{Power-Drill Bit Removal}.
Because these perturbations place the policy in off-nominal contact states, success requires recovery through re-alignment or delayed hand-skill activation rather than replaying a nominal trajectory.

\section{Implementation Details}
\label{appendix:implementation_details}

\subsection{Policy Training Details}
\vspace{-2mm}
Across Sharpa and XHand, coordination policies share the same high-level action interface and RL pipeline.
Task-dependent differences are the active camera streams and online interaction required during fine-tuning, with \textit{Thin-Shaft Tool Retrieval} using the wrist camera and \textit{Pipette Dispensing} using both wrist and side cameras.

Table~\ref{tab:common_hyperparams} summarizes common policy hyperparameters, including control frequency, episode length, network size, optimizer, discount factor, and image augmentation.
\begin{table}[H]
\centering
\small
\renewcommand{\arraystretch}{0.88}
\resizebox{\linewidth}{!}{%
\begin{tabular}{ll|ll}
\toprule
\textbf{Parameter} & \textbf{Value} & \textbf{Parameter} & \textbf{Value} \\ \midrule
Observation space & \begin{tabular}[c]{@{}l@{}}wrist and/or side, tcp\_pose, tcp\_vel, tcp\_f/t \\ hand\_pose, hand\_f/t\end{tabular} & Environment update frequency & 20~Hz \\ 
Action space & 6D twist, skill modulation weight & Proprio encoder size & 64 \\
Reward function & Binary classifier & Policy MLP size & 256$\times$256 \\
Classifier views & wrist and/or side & Discount factor & 0.98 \\
Classifier accuracy & 95\% & Optimizer & Adam \\
Initial offline demonstrations & 100 & Learning rate & 3e{-}4 \\
Offline global BC weight $\lambda_{\text{gbc}}$ & 0.5 & IAW imitation-loss weight $\lambda_I$ & 0.1 \\
Reset method & Scripted/Human reset & Image augmentation & Random crop \\
Max episode length & 2000 environment steps & & \\ \bottomrule
\end{tabular}%
}
\vspace{0.5em}
\caption{Common policy training hyperparameters shared across tasks.}
\label{tab:common_hyperparams}
\end{table}

For hand skill acquisition, kinesthetic prior collection is typically the time needed to record one reference trajectory and usually takes less than five minutes per task.
Because the hand skill reward uses fingertip force feedback, we avoid rapid or jerky contact motions during teaching, which can introduce noisy force and contact measurements.
In practice, we teach the contact sequence slowly to obtain consistent tactile signals, and temporally resample the recorded prior when faster reference execution is desired.
During early residual-RL training, we initialize the residual action near zero and use a small residual action scale so the policy stays close to the kinesthetic prior while collecting initial contact data.
The residual scale is then increased to allow reward-driven adaptation to real contact dynamics.

For arm-hand coordination learning, Table~\ref{tab:training_time_breakdown} reports the wall-clock time spent on reward-classifier data collection, automated DMP rollout collection, and online coordination-policy fine-tuning.
For the reward-classifier column, the reported time includes both positive and negative frame collection.
Across the six task datasets, collection takes $4$--$13$ minutes per task and includes 10--35 positive frames with roughly 1.9k--7.9k negative frames.
The imbalance mainly comes from sampling failure frames from rollout videos, while positive frames are short clips of completed outcomes.
In practice, classifier quality depends on whether the wrist or side camera can observe the task-completion state and distinguish it from near-success failures.
When this visibility condition holds, classifier data collection is inexpensive.
\begin{table}[H]
\centering
\small
\resizebox{\linewidth}{!}{%
\begin{tabular}{l|ccc}
\toprule
\textbf{Task} &
\shortstack{\textbf{Reward Classifier}\\\textbf{Data Collection (mins)}} &
\shortstack{\textbf{DMP Data}\\\textbf{Collection (mins)}} &
\shortstack{\textbf{Coordination Policy}\\\textbf{Training (mins)}} \\
\midrule
Power-Drill Actuation & 4 & 34 & 60 \\
Angle-Spreader Retrieval & 13 & 13 & 55 \\
Thin-Handle Retrieval & 11 & 17 & 60 \\
Power-Drill Bit Removal & 7 & 21 & 22 \\
Pipette Dispensing & 10 & 14 & 60 \\
Thin-Shaft Tool Retrieval & 7.5 & 14 & 20 \\
\bottomrule
\end{tabular}%
}
\vspace{0.5em}
\caption{\textbf{Per-task time cost for coordination learning.}
All times are wall-clock minutes. DMP data collection corresponds to automated rollout collection, and coordination policy training corresponds to online HIL fine-tuning.}
\label{tab:training_time_breakdown}
\end{table}

\subsection{Baseline Details}
\label{appendix:baseline_details}

\noindent\textbf{HIL-SERL variants.}
\rebuttalrev{The two HIL-SERL baselines in Tab.~\ref{tab:main_results} use the same RLPD backbone and offline DMP buffer as \methodname{} but keep a flat policy that directly outputs full arm-hand actions.}
\textit{HIL-SERL (Glove+Gello)} provides full arm-hand corrections using AnyTeleop~\cite{qinAnyTeleopGeneralVisionBased2024} for hand-mapped teleoperation and Gello~\cite{wu2024gello} for arm control.
\textit{HIL-SERL (Ours Interv.)} uses the same learned hand-skill intervention interface as \methodname{} while retaining the flat full-DoF policy.
This separates the effect of the intervention interface from the effect of the hierarchical architecture.

\noindent\textbf{Behavior Cloning.}
BC is trained on the same 200 human demonstrations and uses the same frozen pre-trained ResNet-10 visual encoder as our method for fair comparison.
Visual features from each camera view and proprioceptive inputs are concatenated and processed by an MLP that outputs a diagonal Gaussian policy.
Training minimizes the negative log-likelihood loss $\mathcal{L}_{\text{BC}}=-\mathbb{E}[\log \pi_\theta(a\mid o)]$ using Adam with learning rate $3\times10^{-4}$ and batch size 256 for 20,000 steps.
Random cropping is applied as image augmentation, and zero-action transitions are filtered from demonstrations.

\noindent\textbf{Diffusion Policy.}
DP~\cite{chi2025diffusion} is trained on the same demonstrations as BC and uses a 1D temporal U-Net noise prediction network, with fully fine-tuned ViT-B/16~\cite{Dosovitskiy2020AnII} encoders for each camera view without weight sharing.
The observation and action prediction horizons are set to 2 and 16, respectively.
Training uses DDIM with 50 diffusion steps and a squared cosine noise schedule, optimizing the MSE loss on predicted noise with AdamW (learning rate $1\times10^{-4}$, batch size 256) for 100 epochs.
EMA with decay 0.995 is applied.

\noindent\textbf{Additional IL training protocol.}
The main BC/DP comparison tests whether offline imitation from successful human demonstrations is sufficient for functional dexterous tasks.
To separate data coverage from the learning paradigm, we additionally train BC and DP on successful trajectories collected during \methodname{} training, including DMP rollouts, online policy rollouts, and task-completing human-intervention segments.
Additional results for these same-source IL policies are reported in Sec.~\ref{appendix:additional_il_results}.

\raggedbottom
\section{Additional Experimental Results}
\label{appendix:additional_analysis}
\vspace{-0.8em}

\rebuttalrev{\noindent\textbf{Project videos.}}
\rebuttalrev{The \href{https://hly-123.github.io/HiFun/}{project website} shows all six tasks, position and fingertip-contact recovery, tool-size transfer, and a long-horizon multi-skill sequence.}

\subsection{Online Training Curves and Autonomy}
\label{appendix:learning_curves}
\vspace{-0.2em}

\begin{figure}[!t]
\centering
\includegraphics[
    page=1,
    width=\linewidth,
    height=0.66\textheight,
    keepaspectratio
]{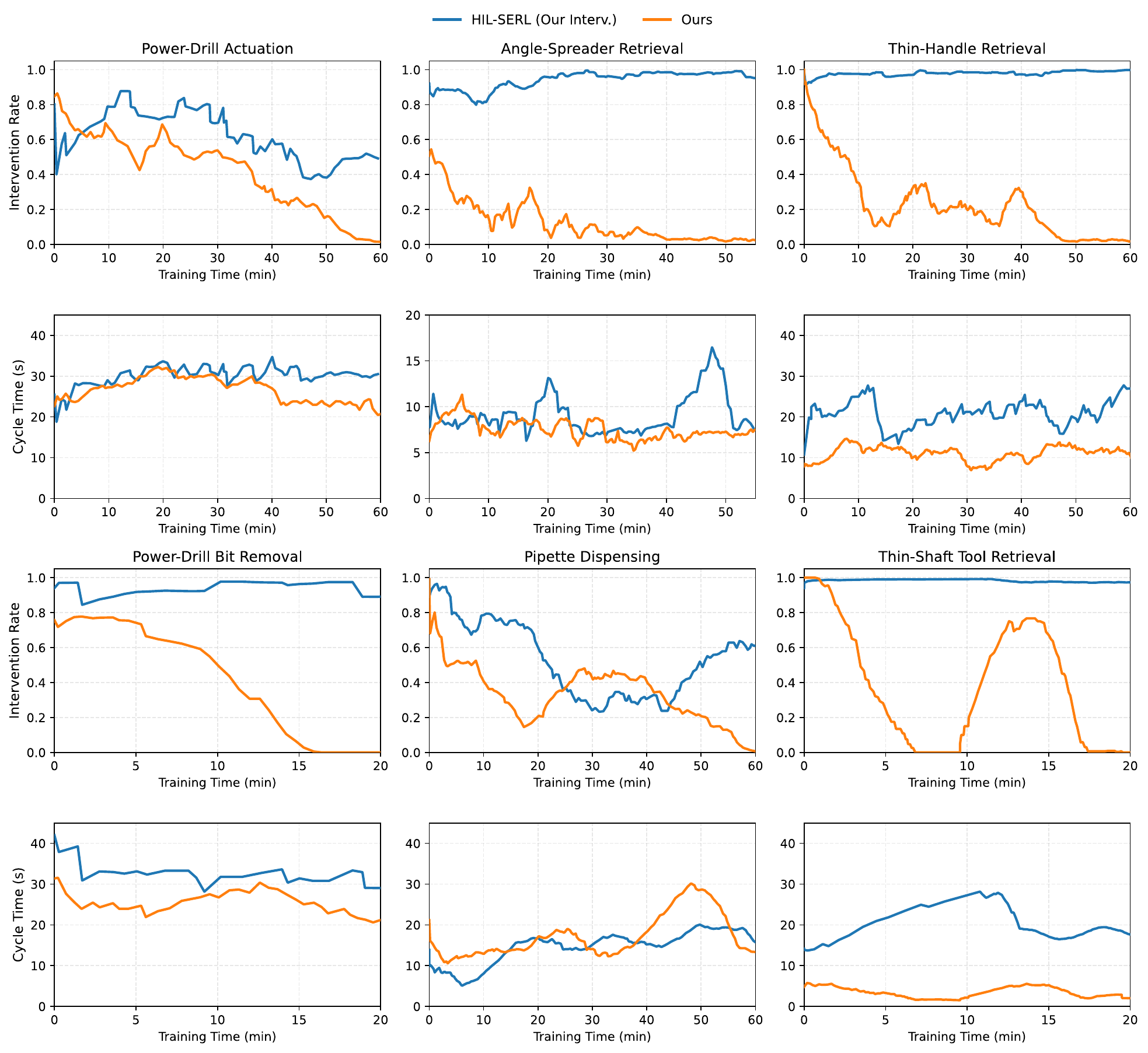}
\caption{\textbf{Online training curves across six tasks.}
For each task, intervention rate (upper) and cycle time (lower) are shown as running averages over 20 episodes, comparing our method (Ours) against the HIL-SERL (Ours Interv.) baseline.
Lower intervention rate indicates more autonomous execution, while lower cycle time indicates more efficient completion.}
\label{fig:learning_curves}
\vspace{0.5em}
\centering
\small
\resizebox{0.9\linewidth}{!}{%
\begin{tabular}{l|cc|cc}
\toprule
\multirow{2}{*}{\textbf{Task}} &
\multicolumn{2}{c|}{\textbf{BC Success Rate (\%)}} &
\multicolumn{2}{c}{\textbf{DP Success Rate (\%)}} \\
\cmidrule(lr){2-3}\cmidrule(lr){4-5}
& \textbf{Human Demos} & \textbf{\methodname{} Data}
& \textbf{Human Demos} & \textbf{\methodname{} Data} \\
\midrule
Power-Drill Actuation & 15 & 12 & 30 & 26 \\
Angle-Spreader Retrieval & 30 & 32 & 60 & 64 \\
Thin-Handle Retrieval & 35 & 32 & 40 & 40 \\
Power-Drill Bit Removal & 30 & 28 & 36 & 40 \\
\midrule
Average & 27.5 & 26.0 & 41.5 & 42.5 \\
\bottomrule
\end{tabular}%
}
\captionof{table}{\textbf{Data-source comparison for IL baselines.}
Success rates follow the main evaluation protocol on the Sharpa tasks. Human Demos uses the original 200 teleoperated demonstrations, while \methodname{} Data uses successful trajectories collected during \methodname{} training.}
\label{tab:il_hifun_data}
\end{figure}

Figure~\ref{fig:learning_curves} reports intervention rate and cycle time during online training across the six tasks for \methodname{} and HIL-SERL (Ours Interv.).
Since each successful human correction completes the task during training, success rate alone does not reveal whether the learned policy becomes autonomous.
The intervention rate therefore measures how often human guidance is still needed, and cycle time measures execution efficiency.
For \methodname{}, the intervention rate decreases toward zero within one hour of online training on every task (and in as little as ${\sim}15$ minutes on the fastest tasks), and its cycle time at convergence is consistently lower than HIL-SERL (Ours Interv.).
In contrast, the baseline keeps a high intervention rate and never becomes autonomous within the same one-hour training budget, despite using the same intervention interface.
This supports the role of \methodname{} in efficient online learning.

\FloatBarrier

\subsection{Additional IL Baseline Results}
\label{appendix:additional_il_results}

Table~\ref{tab:il_hifun_data} shows that replacing human demonstrations with \methodname{} training trajectories does not materially improve offline imitation: BC changes from 27.5\% to 26.0\% average success, and DP changes from 41.5\% to 42.5\%.
Policies trained on \methodname{} data often approach objects faster and hesitate less near intermediate states, consistent with exposure to rollouts and intervention segments from more varied contact states.
However, they still fail to reliably complete precise alignment and contact recovery.
BC and DP are trained by supervised action prediction over the collected trajectories, which does not provide a value-based mechanism for selecting which recovery-like actions should be reinforced for downstream task completion.
This explains why recovery-rich offline data alone does not yield reliable recovery behavior without value-guided online RL.

\begingroup
\rebuttalcolor
\subsection{Skill-HIL Comparison}
\label{appendix:skill_hil}

We examine whether providing a learned hand skill to standard HIL-SERL is sufficient to reproduce \methodname{}'s performance.
The skill reduces the burden of exploring and correcting high-dimensional finger actions, but two challenges remain in coordination learning.
First, sparse task outcomes do not directly distinguish errors in arm alignment from errors in skill activation, making credit assignment between these action components difficult.
Second, successful human interventions differ in recovery efficiency and consistency, and therefore provide supervision of varying usefulness.
\methodname{} addresses these challenges with separate value learning for arm alignment and skill activation, plus intervention advantage weighting that scores each correction against the policy action.

To test whether HiFun's gains come from giving HIL-SERL the learned hand skill, we evaluate \textbf{Skill-HIL}, which combines that frozen skill with standard HIL-SERL coordination learning.
Across four Sharpa tasks under matched settings and training budgets, Skill-HIL achieves 46\% average success, compared with 100\% for \methodname{}.
Most failures involve incorrect contact regions or activation timing, followed by collisions, consistent with unresolved coordination errors.

On \textit{Angle-Spreader Retrieval}, doubling Skill-HIL's training data yields 35/50 successful trials, compared with 50/50 for \methodname{}.
Thus, Skill-HIL's average remains lower across the four tasks, and doubling its data does not close the gap on \textit{Angle-Spreader Retrieval}.
This comparison evaluates HiFun's coordination design as a whole; Tab.~\ref{tab:high_level_analysis} separately ablates the multi-head critic and intervention advantage weighting.

\endgroup

\subsection{Coordination Component Analysis}
\label{appendix:value_guided}
\begin{figure}[H]
    \centering
    \includegraphics[width=\linewidth]{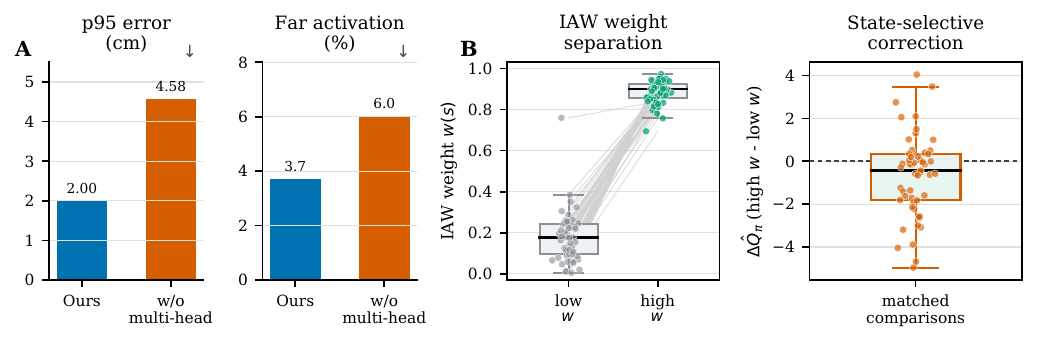}
    \caption{\textbf{Coordination component analysis.}
    (A) Activation localization on \textit{Angle-Spreader Retrieval}: the x-axis compares \methodname{} with the one-head-critic variant, and the y-axes report p95 activation error and far-state activation rate computed from TCP-to-target error $e_t$ with a $2\,\mathrm{cm}$ pre-contact band.
    (B) IAW weighting: left shows imitation weight $w(s)$ for low- and high-weight interventions; right shows $\Delta \hat Q_\pi=\hat Q(s_{\mathrm{high\_w}},\bar a_\pi(s_{\mathrm{high\_w}}))-\hat Q(s_{\mathrm{low\_w}},\bar a_\pi(s_{\mathrm{low\_w}}))$, where negative values indicate high-weight corrections at lower-value policy states.}
    \label{fig:value_guided_coordination_appendix}
\end{figure}

Figure~\ref{fig:value_guided_coordination_appendix} visualizes the mechanisms behind the coordination ablations in the main paper.
In Fig.~\ref{fig:value_guided_coordination_appendix}A, the multi-head critic reduces the 95th-percentile activation error from $4.58\,\mathrm{cm}$ to $2.00\,\mathrm{cm}$ and reduces activations outside the $2\,\mathrm{cm}$ pre-contact band from $6.0\%$ to $3.7\%$.
This indicates that separating arm-alignment and skill-activation value learning localizes hand-skill activation closer to the target region, supporting better temporal coordination between arm approach and hand activation.
In Fig.~\ref{fig:value_guided_coordination_appendix}B, IAW assigns clearly separated imitation weights to weak and strong corrections ($0.175$ vs. $0.884$ mean).
The matched policy-value comparison is negative, showing that high-weight corrections occur at states where the current policy action has lower estimated value.
This supports the main-text claim that IAW avoids uniform cloning by focusing imitation on useful corrections while limiting low-advantage corrections that could conflict with critic optimization.

\subsection{Hand Motion Control Failure Cases}
\label{appendix:hand_failure}
\begin{figure}[H]
    \centering
    \includegraphics[page=1,width=\linewidth]{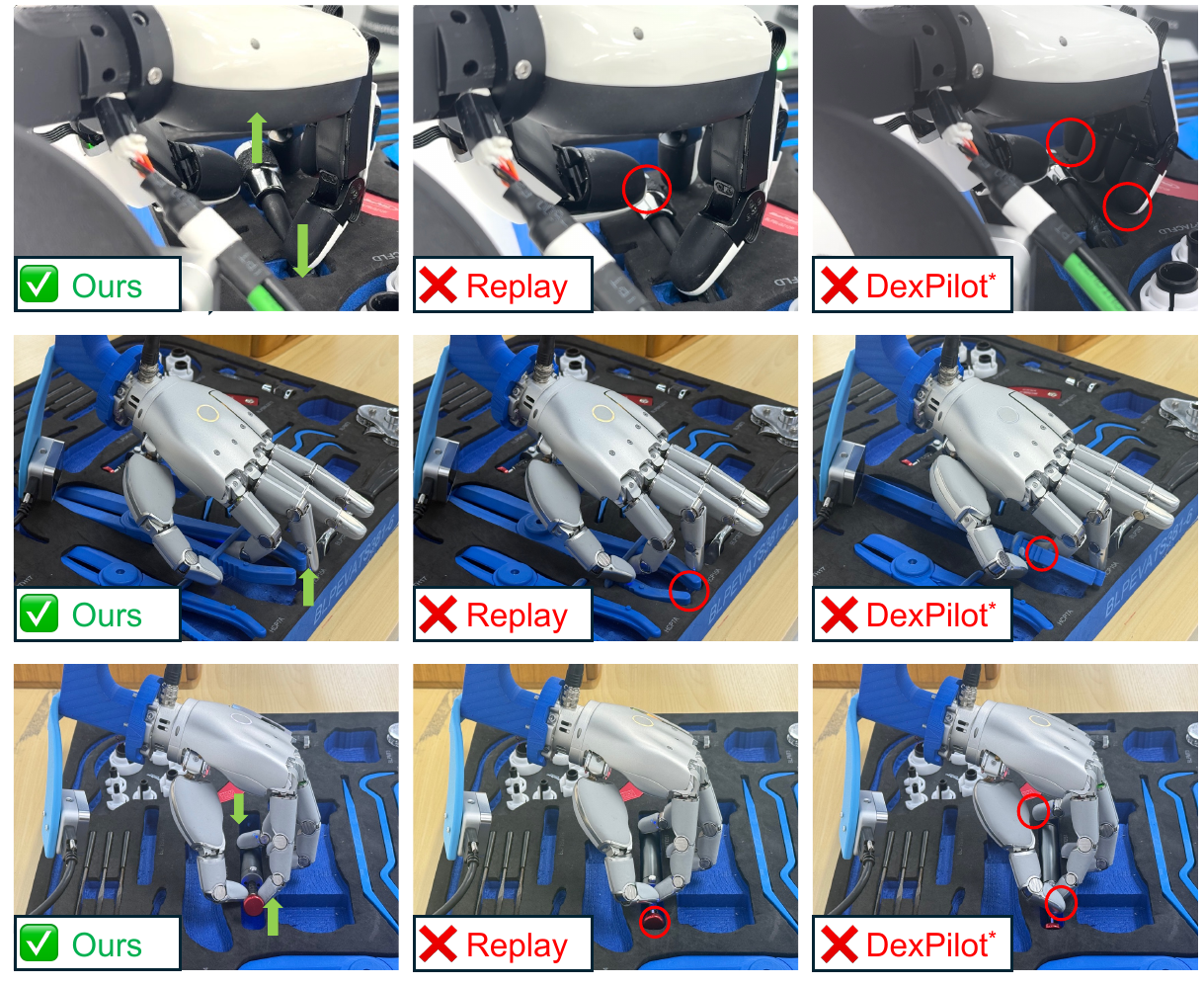}
    \caption{\textbf{Failure analysis for hand motion control.}
    Representative hand-control outcomes compare the residual hand skill, kinesthetic replay, and DexPilot$^*$ across contact-rich manipulation cases.}
    \label{fig:hand_skill_failure}
\end{figure}
Figure~\ref{fig:hand_skill_failure} compares hand-control methods on three contact-rich tasks: \textit{Thin-Shaft Tool Retrieval} with XHand, \textit{Angle-Spreader Retrieval} with Sharpa, and \textit{Thin-Handle Retrieval} with Sharpa.
After placing the arm near the task execution region and focusing on the task-relevant fingers, the failures mainly reflect functional finger execution rather than arm placement or full-hand teleoperation noise.
Kinesthetic replay preserves the demonstrated motion pattern but executes it open-loop, losing the contact-force information needed to lift or lever objects with the task-relevant fingers.
DexPilot$^*$, despite fingertip-based retargeting and per-task tuning, can produce coarse task-relevant contacts but remains sensitive to the human-robot morphology gap in constrained settings: in \textit{Angle-Spreader Retrieval}, it can lift the tool but forms an unstable grasp, while in \textit{Thin-Shaft Tool Retrieval} and \textit{Thin-Handle Retrieval}, it struggles with precise pinch alignment and little-finger levering.
In contrast, our residual hand skill adapts the kinesthetic prior with real contact feedback, enabling closed-loop force modulation for reliable functional hand execution.

\end{document}